%% file: main.tex
\documentclass[conference]{IEEEtran}
\IEEEoverridecommandlockouts
\usepackage{cite}
\usepackage{amsmath,amssymb,amsfonts}
\usepackage{graphicx}
\usepackage{textcomp}
\usepackage{multirow}
\usepackage{xcolor}
\usepackage{comment}
\usepackage{float}
\usepackage{caption}
\usepackage{threeparttable}
\usepackage{subcaption}
\usepackage{mathtools}
\usepackage{todonotes}
\usepackage[mathscr]{euscript}
\usepackage{url}
\usepackage{hyperref}
\usepackage{bm}
\usepackage{algpseudocode} 
\usepackage{algorithm}
\usepackage{xifthen}
\usepackage[export]{adjustbox}
\usepackage{textcomp}
\usepackage{array}   
\usepackage{svg}
\usepackage{enumitem}
\usepackage{colortbl}
\newcommand{\mycomment}[1]{}
\NewDocumentCommand{\vect}{ O{} O{} m }{\mathbf{#3}\ifthenelse{\isempty{#1}}{}{^{(#1)}}\ifthenelse{\isempty{#2}}{}{_{#2}}}
\NewDocumentCommand{\mat}{ O{} O{} m }{\mathbf{#3}\ifthenelse{\isempty{#1}}{}{^{(#1)}}\ifthenelse{\isempty{#2}}{}{_{#2}}}
\NewDocumentCommand{\ten}{ O{} O{} m }{\pmb{\mathscr{#3}}\ifthenelse{\isempty{#1}}{}{^{(#1)}}\ifthenelse{\isempty{#2}}{}{_{#2}}}

\def\BibTeX{{\rm B\kern-.05em{\sc i\kern-.025em b}\kern-.08em
    T\kern-.1667em\lower.7ex\hbox{E}\kern-.125emX}}

\definecolor{mygreen}{rgb}{0,0.6,0}
\definecolor{mygray}{rgb}{0.5,0.5,0.5}
\definecolor{mymauve}{rgb}{0.58,0,0.82}

\usepackage{listings}
\lstdefinelanguage{cypher}{
    sensitive=true,
    morekeywords=[1]{MATCH, RETURN, WHERE},
    morekeywords=[2]{PERSON, FRIEND},
    morestring=[b]",
    morecomment=[l]{//},
    morecomment=[s]{/*}{*/},
    morecomment=[s]{--}{\ },
}

\lstdefinestyle{cypherstyle}{
    language=cypher,
    basicstyle=\small\ttfamily,
    keywordstyle=[1]\color{blue},
    keywordstyle=[2]\color{red},
    commentstyle=\color{mygreen},
    stringstyle=\color{mymauve},
    numberstyle=\tiny\color{mygray},
    breaklines=true,
    showstringspaces=false,
    captionpos=b
}

\begin{document}

\title{Tensor Train Low-rank Approximation (TT-LoRA): Democratizing AI with Accelerated LLMs}

\author{\IEEEauthorblockN{
 Afia Anjum\IEEEauthorrefmark{1}\IEEEauthorrefmark{3},
 Maksim E. Eren\IEEEauthorrefmark{2}, 
 Ismael Boureima\IEEEauthorrefmark{1},
 Boian Alexandrov\IEEEauthorrefmark{1},
 Manish Bhattarai\IEEEauthorrefmark{1},
 }
 \IEEEauthorblockA{
 \IEEEauthorrefmark{1}Theoretical Division, Los Alamos National Laboratory. Los Alamos, USA. \\
 \IEEEauthorrefmark{2}Advanced Research in Cyber Systems, Los Alamos National Laboratory. Los Alamos, USA. \\
 \IEEEauthorrefmark{3}University of Texas at Arlington. Texas, USA. \\}

 \thanks{U.S. Government work not protected by U.S. copyright.} }

\maketitle
\vspace{-40em}

\begin{abstract}
In recent years, Large Language Models (LLMs) have demonstrated remarkable capabilities across a wide range of natural language processing (NLP) tasks, such as question-answering, sentiment analysis, text summarization, and machine translation. However, the ever-growing complexity of LLMs demands immense computational resources, hindering the broader research and application of these models. To address this, various parameter-efficient fine-tuning strategies, such as Low-Rank Approximation (LoRA) and Adapters, have been developed. Despite their potential, these methods often face limitations in compressibility. Specifically, LoRA struggles to scale effectively with the increasing number of trainable parameters in modern large scale LLMs. Additionally, Low-Rank Economic Tensor-Train Adaptation (LoRETTA), which utilizes tensor train decomposition, has not yet achieved the level of compression necessary for fine-tuning very large scale models with limited resources. This paper introduces Tensor Train Low-Rank Approximation (TT-LoRA), a novel parameter-efficient fine-tuning (PEFT) approach that extends LoRETTA with optimized tensor train (TT) decomposition integration. By eliminating Adapters and traditional LoRA-based structures, TT-LoRA achieves greater model compression without compromising downstream task performance, along with reduced inference latency and computational overhead. We conduct an exhaustive parameter search to establish benchmarks that highlight the trade-off between model compression and performance. Our results demonstrate significant compression of LLMs while maintaining comparable performance to larger models, facilitating their deployment on resource-constraint platforms. 
\end{abstract}


\section{Introduction}
\label{sec:introduction}
\input{00-sec_introduction}

\section{Related Works}
\label{sec:related_works}

\input{01-sec_relatedwork}

\section{Tensor Train based Low-Rank Adaptation (TT-LoRA)}
\label{sec:methods}
\input{02-sec_method}
 
 \section{Results and Discussion}
\label{sec:results}
\input{03-sec_results}

\section{Conclusion}
\label{sec:conclusion}
\input{04-sec_conclusion}

\bibliographystyle{IEEEtran}
\bibliography{main}

\vspace{12pt}

\end{document}

%% file: 00-sec_introduction.tex
\begin{figure}[ht!]
    \centering
    \includegraphics[width=0.4\textwidth]{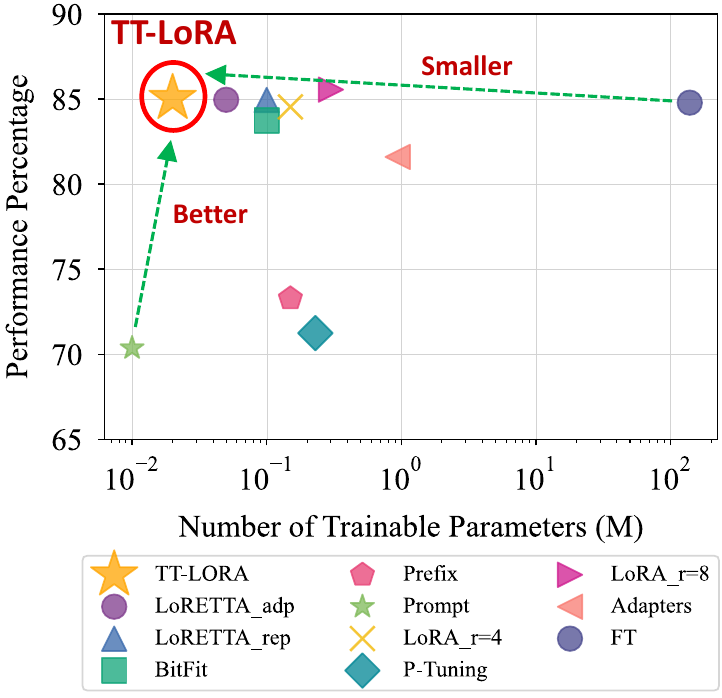}
    \caption{Performance vs. trainable parameters comparison between TT-LoRA and various PEFT methods on DeBERTa model using the GLUE benchmark}
    \label{fig:comparison}
    \vspace{-5mm}
\end{figure}

Large Language Models (LLMs) such as LLaMA-70B\cite{1}, ChatGPT-4\cite{2}, Bard\cite{3}, and Claude\cite{3} represent a significant step towards Artificial General Intelligence (AGI)\cite{4}. These models are trained using vast amounts of data and are complex neural network architectures, such as Transformers~\cite{15}, enabling the models to excel in interpreting complex linguistic patterns. Consequently, these models can accurately recognize, translate, predict, and generate text, along with performing other content-related tasks. The effectiveness of LLMs in capturing the nuances of language has made them indispensable tools for driving previously unattainable innovations.

While pre-trained LLMs provide a robust foundation for general language tasks, fine-tuning these models on application-specific datasets is crucial for optimizing performance in specialized applications, which allows the models to adapt to a particular context or domain. However, fine-tuning involves adapting all the parameters of a pre-trained model to a new tasks, which poses significant challenges due to the reliance of these models on increasingly complex Transformers with exploding parameter counts (model size in billions of parameters), requiring substantial computational resources~\cite{5}. Fine-tuning these large models through traditional approaches, which involves scaling these models using multiple GPUs, is hindered by resource limitations and rising costs, monopolizing research access and raising ethical concerns. Additionally, the immense computational demands of these models raise serious environmental concerns due to the energy consumption of massive computing complexes \cite{6}. 

Since full model fine-tuning, which involves adjusting all the parameters of a pre-trained model, becomes prohibitively expensive as the model size of LLMs grows, notable pragmatic efforts, such as parameter-efficient fine tuning (PEFT) techniques like Adapters\cite{7}, Prefix Tuning~\cite{li2021prefix}, Prompt Tuning~\cite{lester2021power}, Low-Rank Adaptation (LoRA)\cite{8}, and Low-Rank Economic Tensor-Train Adaptation (LoRETTA)~\cite{yang2024loretta} have been proposed for efficient LLM fine-tuning. However, Adapter-based PEFT methods increase model's inference latency while Prompt and Prefix Tuning sacrifice model accuracy. Moreover, when scaling to recently proposed larger models such as LAaMA3-70B and Mixture of Experts (MoE), the number of trainable parameters needed in LoRA and LoRETTA methods is still high, limiting these models' scalability and application. 

To address the scalability issues of PEFT methods with increasing model parameters, in this paper, we propose Tensor Train Low Rank Approximation (TT-LoRA), a novel PEFT approach using Tensor Train (TT) decomposition~\cite{oseledets2011tensor}. TT decomposition has already shown promise in compressing and accelerating neural networks by efficiently representing large weight matrices (model parameters) in a compact tensor format, facilitating substantial reductions in computational load without severely compromising performance\cite{13, 14}. In addition, the LoRETTA approach has demonstrated the efficacy of utilizing tensor train (TT) decomposition for refining weight updates, achieving notable accuracy enhancements in LLM applications. Motivated by the efficiency of TT decomposition, we explore its potential and extend LoRETTA with architectural variations. Our approach, TT-LoRA, diverges significantly by optimizing the integration process of TT decomposition into the model’s architecture, specifically, omitting Adapters and the LoRA-based structure employed in LoRETTA. This crucial modification eliminates the additional inference latency associated with these elements, achieves superior model compression, and reduces model complexity and computation overhead. 

Our contributions are:

\begin{itemize}
    \item We introduce Tensor Train Low Rank Approximation (TT-LoRA), a parameter-efficient fine-tuning strategy for large language models (LLMs). TT-LoRA leverages tensor train decomposition to enable fine-tuning of LLMs while significantly reducing the number of trainable parameters.
    
    \item We conduct a comprehensive evaluation of TT-LoRA's performance across a variety of downstream tasks and LLMs of differing scales, including BERT-based models and the larger-scale LLaMA-2 and LLaMA-3 models. 

    \item We benchmark the performance of TT-LoRA against other widely PEFT methods across diverse model scales. Our results show that TT-LoRA achieves greater performance while ensuring significant model compression, as shown in Figure~\ref{fig:comparison}.

    \item We perform an exhaustive parameter search to establish benchmarks that illustrate the trade-offs between model compression and performance, providing valuable insights for optimizing PEFT methods.
    
\end{itemize}

%% file: 01-sec_relatedwork.tex
\begin{figure*}[ht!]
\centering
\begin{subfigure}{.27\textwidth}
  \centering
  \includegraphics[width=\linewidth]{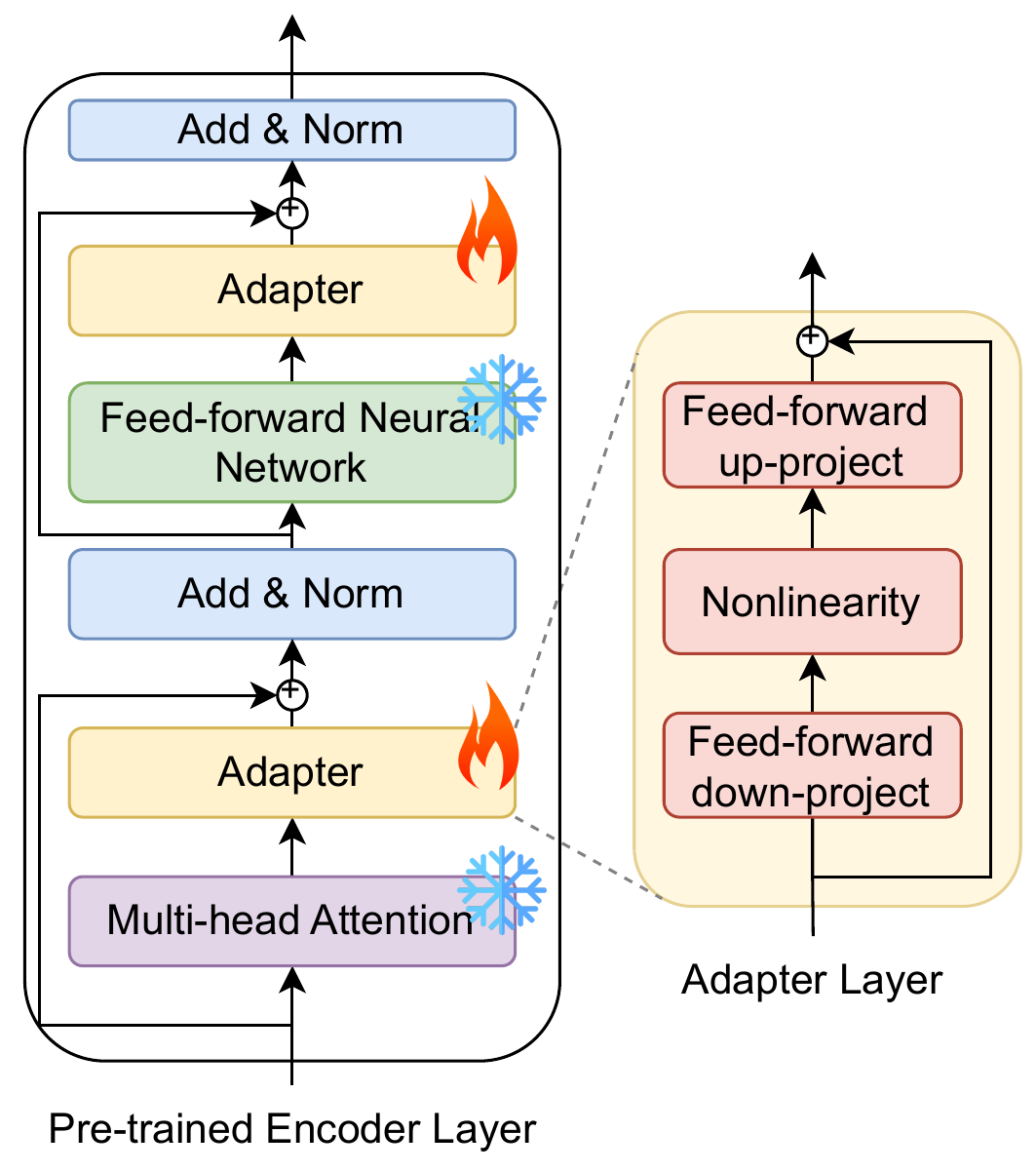}
  \caption{Adapters}
  \label{fig:adp}
\end{subfigure}\quad
\begin{subfigure}{.23\textwidth}
  \centering
  \includegraphics[width=\linewidth]{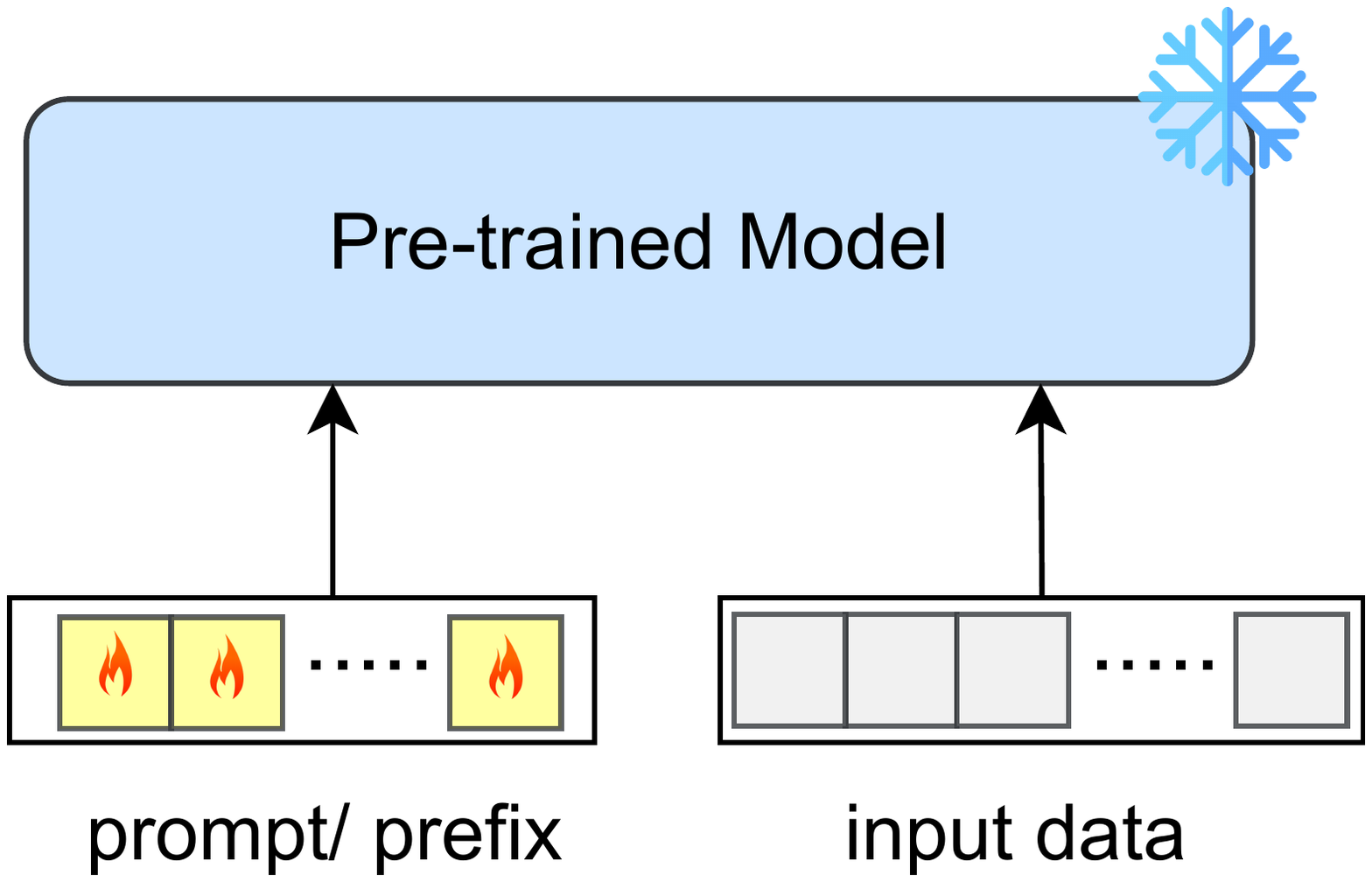}
  \caption{Prompt, Prefix, and P-Tuning}
  \label{fig:tuning}
\end{subfigure}\quad
\begin{subfigure}{.37\textwidth}
  \centering
  \includegraphics[width=\linewidth]{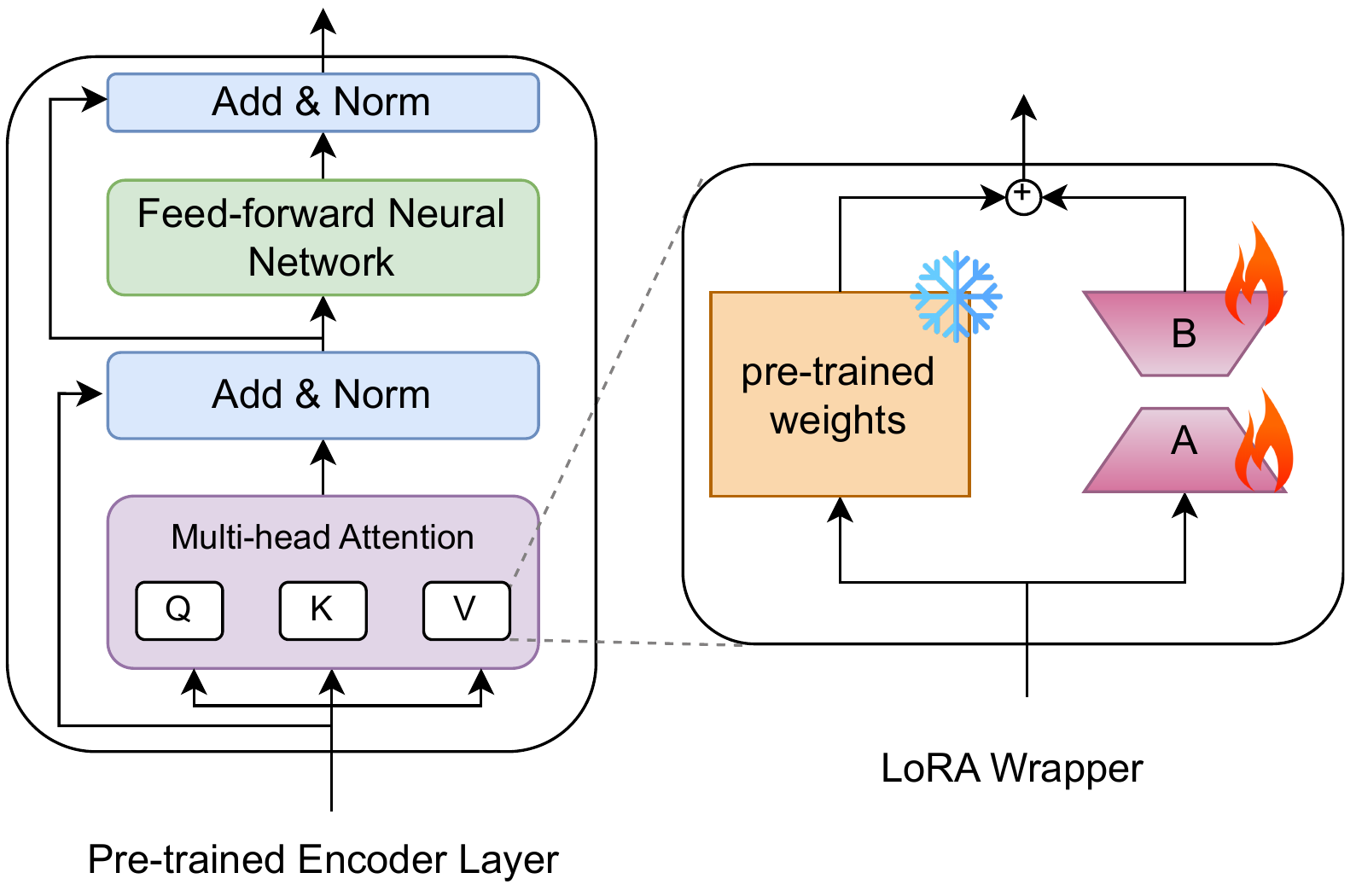}
  \caption{LoRA}
  \label{fig:lora}
\end{subfigure}

\medskip

\begin{subfigure}{.27\textwidth}
  \centering
  \includegraphics[width=\linewidth]{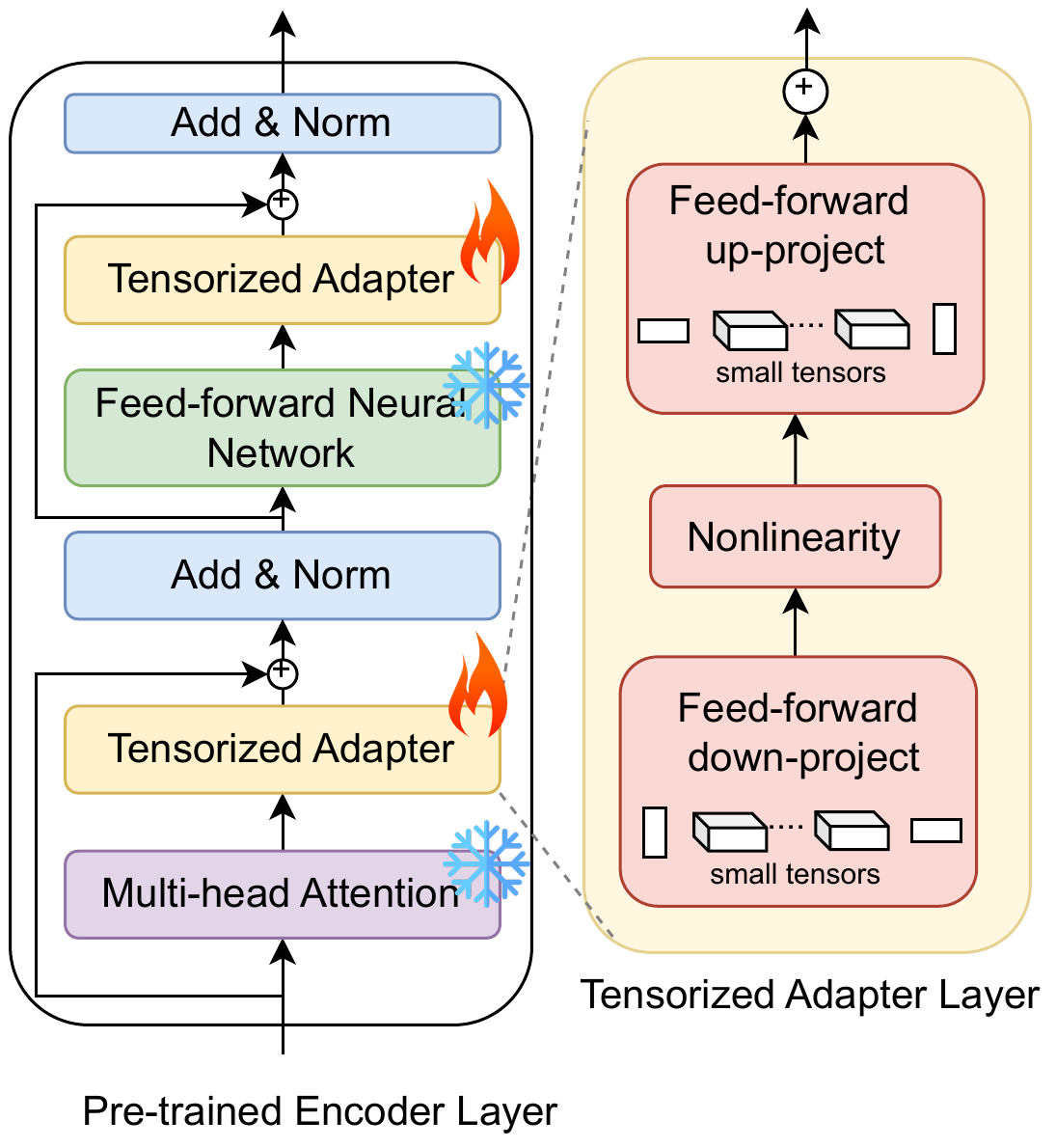}
  \caption{LoRETTA$_{adp}$}
  \label{fig:loreta-adp}
\end{subfigure}\quad
\begin{subfigure}{.63\textwidth}
  \centering
  \includegraphics[width=\linewidth]{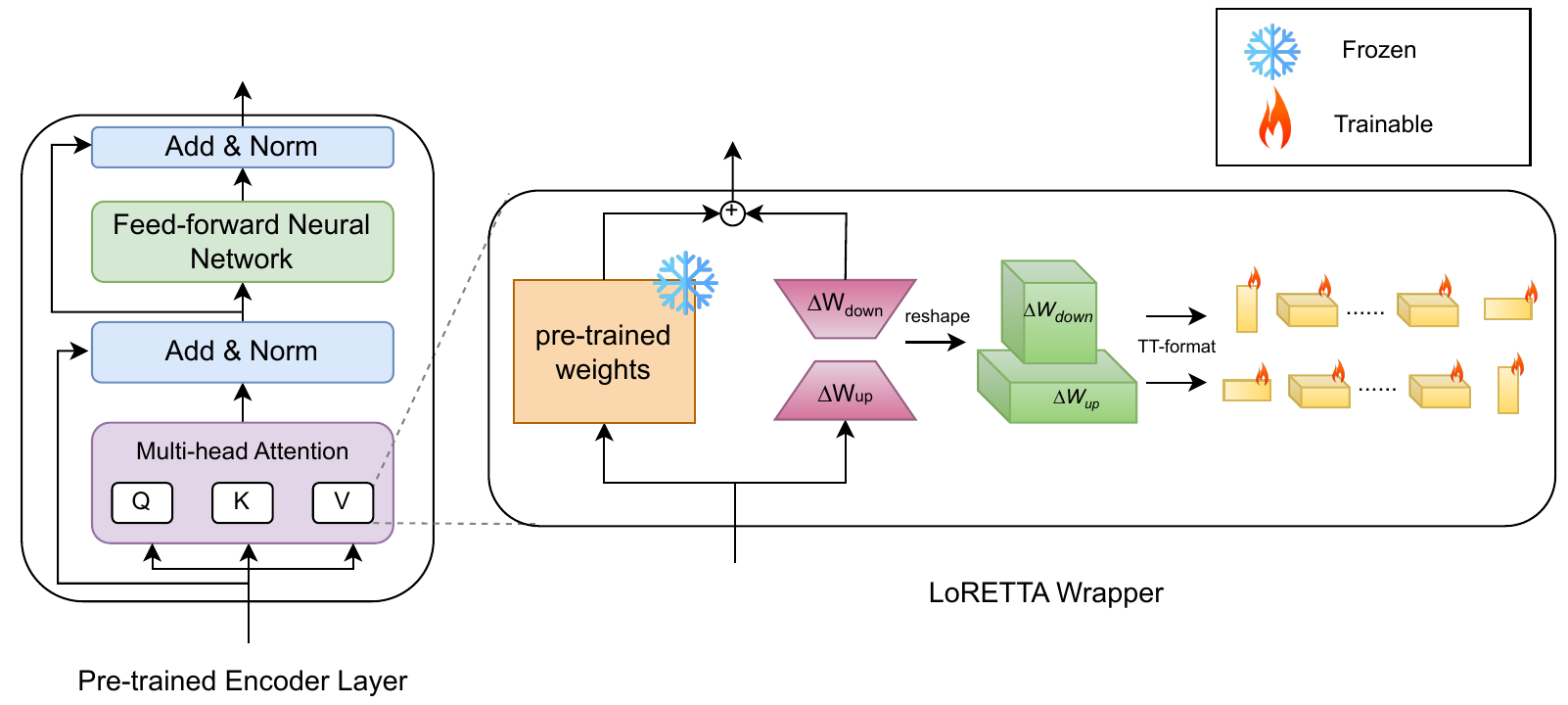}
  \caption{LoRETTA$_{rep}$}
  \label{fig:loreta-rep}
\end{subfigure}

\caption{Various Parameter-efficient Fine-Tuning Approaches}
\label{fig:rel-works}
    \vspace{-5mm}
\end{figure*}

In this section, we explore various strategies for parameter-efficient fine-tuning of LLMs, as illustrated in Figure~\ref{fig:rel-works}, which are crucial for deploying these models in resource-constrained environments.

\textbf{Adapter-based Methods:}
Adapter-based methods~\cite{houlsby2019parameter, lin2020exploring, pfeiffer2020adapterfusion} introduce small, trainable modules within the pre-trained model's architecture, where each module consists of fully connected layers configured with a bottleneck structure (Figure~\ref{fig:adp}). keeping most of the model's parameters unchanged. This adapter-based design keeps the pre-trained model's weights fixed while only updating the parameters within the adapters during task-specific fine-tuning. Although adapter-based methods can reduce the number of trainable parameters, they introduce additional computational steps in the Transformer blocks. Due to their sequential processing nature, these adapter layers do not effectively leverage hardware parallelism, which results in increased latency, particularly in online inference scenarios with small batch sizes~\cite{hu2021lora}.

\textbf{Prefix Tuning, Prompt Tuning and P-tuning:} Li and Liang~\cite{li2021prefix} introduced prefix-tuning, a method where a sequence of continuous task-specific vectors, also known as prefix, is prepended to the model input, optimizing only the prefix while keeping the model parameters frozen. Lester et al.~\cite{lester2021power} simplify the prefix-tuning by proposing prompt-tuning, where $k$ tunable tokens per downstream tasks are prepended to the input text. Liu et al.~\cite{liu2021p} proposed p-tuning, which extends prompt tuning by integrating continuous token embeddings not only at the input level but also at various points throughout the model, enhancing the adjustment of model processing. These tuning methods are illustrated in Figure~\ref{fig:tuning}.
However, these tuning methods occupy a part of the fixed sequence length that transformer-based architectures can process. Consequently, this reduces the available space for actual task-related input, potentially compromising the model's efficiency~\cite{hu2021lora}. 

\textbf{Low-rank Approximation:} Hu et al.~\cite{hu2021lora} proposed a Low-rank Adaptation (LoRA) fine-tuning approach leveraging matrix factorization, which decomposes a large matrix into a product of two or more smaller matrices (Figure~\ref{fig:lora}). For a pre-trained weight matrix $W_0 \in \mathbb{R}^{m \times n}$, the weight update in a full fine-tuning setting would be $W_0 + \Delta W$, where $\Delta W \in \mathbb{R}^{m \times n}$. In contrast, LoRA enables low-rank updates to the weight matrix as $W_0 + BA$, where $W_0$ is kept frozen while only optimizing $B$ and $A$ matrices. Here, $BA$ is the low-rank approximation of $\Delta W$, where $B \in \mathbb{R}^{m \times r}$, $A \in \mathbb{R}^{r \times n}$, and $r \ll min (m, n)$. While LoRA achieves similar or even better performance than full-model fine-tuning, it still incurs a large number of trainable parameters. For instance, when fine-tuning the LLaMA-2-70B model using LoRA, over 16 million parameters need to be updated, exceeding the total number of parameters in some BERT models~\cite{yang2024loretta}.

\textbf{Tensor-based Model Compression:} Yang et al.~\cite{yang2024loretta} proposed Low-Rank Economic Tensor-Train Adaptation (LoRETTA), inspired by the Tensor Train (TT) format initially explored by Novikov et al.~\cite{novikov2015tensorizing}, which represents a matrix with a series of tensor factors. The authors proposed two methods, LoRETTA$_{adp}$ and LoRETTA$_{rep}$. The former method, LoRETTA$_{adp}$, employs tensorized adapters, compressing the weight updating matrix using two tensorized linear layers (Figure~\ref{fig:loreta-adp}). The latter method, LoRETTA$_{rep}$, performs matrix factorization to reduce the large updating matrix into two small matrices, followed by reshaping the two updating matrices into small tensor factors (Figure~\ref{fig:loreta-rep}). 
However, when scaling to recent larger models such as LLaMA2-70B or attempting to leverage the full potential of techniques like Mixture of Experts (MoE), which are inherently resource-intensive, LoRETTA still incurs a large number of trainable parameters.

Our approach, TT-LoRA, employs a comprehensive parameter search to optimize weight updating matrix decomposition. This process involves identifying the optimal tensor shapes and hyperparameters to achieve the most efficient compressed representation with higher accuracy. Unlike LoRETTA, which employs Adapters and the LoRA-format, TT-LoRA directly decomposes the weight updating matrix into small tensors. This approach eliminates the potential for increased inference latency associated with Adapters and achieves more effective compression by avoiding the LoRA-based structure. 
Extensive evaluations have demonstrated that TT-LoRA surpasses LoRETTA's performance on BERT and LLaMA, both on accuracy and model compression, across various classification tasks, establishing it as the superior method for efficient and accurate model representation.

%% file: 02-sec_method.tex
\begin{figure*}[ht!]
    \centering
    \includegraphics[width=0.9\textwidth]{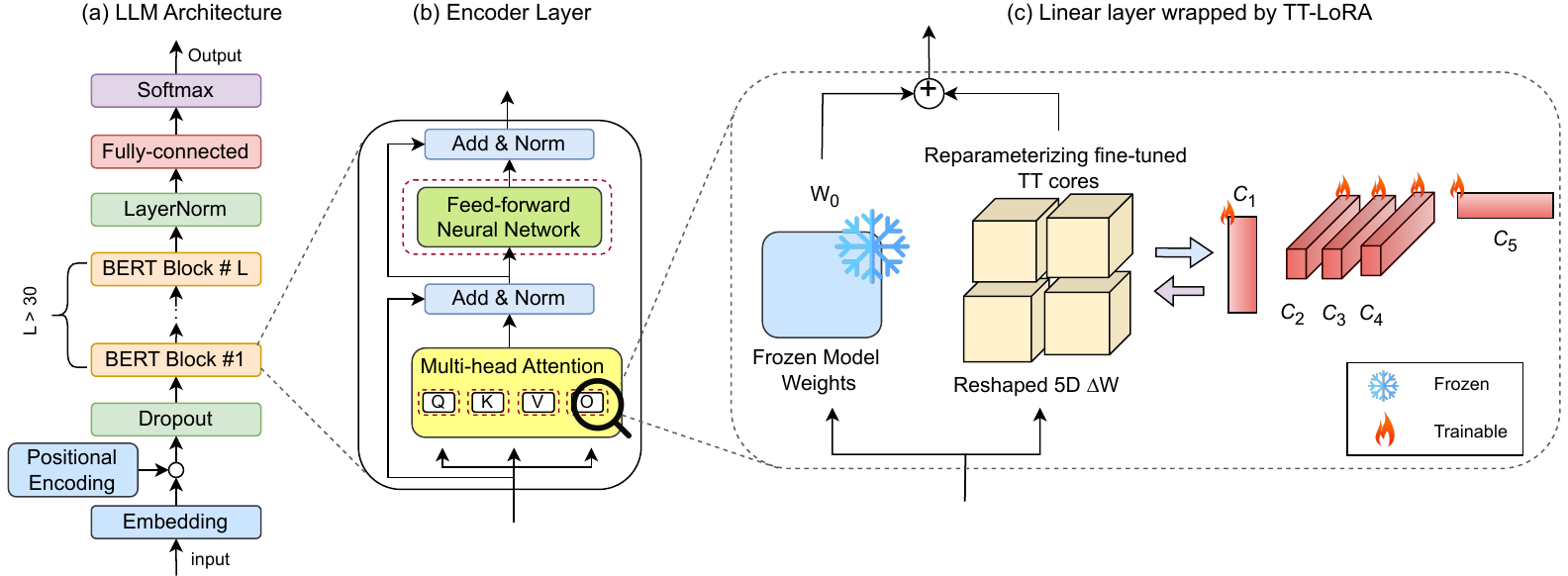}
    \caption{Overview of the proposed TT-Lora framework: (a) BERT Architecture, (b) Encoder layer of BERT Architecture, (c) TT-LoRA wrapping the \textit{value} vector of BERT encoder: the pre-trained weight, $W_0$, is kept frozen while the update to the pre-trained weight, $\Delta W$, is decomposed into small tensors.}
    \label{fig:ttlora_architecture}
    \vspace{-5mm}
\end{figure*}

In this section, we first formulate the objective function of fine-tuning an LLM and outline the challenges associated with conventional full fine-tuning approach. Subsequently, we introduce our proposed PEFT method TT-LoRA, detailing its design and how it addresses these challenges.

\subsection{Problem Statement}

Let $P_\theta (y,x)$ be a pre-trained language model, parameterized by $\theta$. For example, $P_\theta (y,x)$ could represent a versatile multi-task learning model, such as DistilBERT~\cite{sanh2019distilbert}, DeBERTa~\cite{he2020deberta}, or LLaMA-3~\cite{1}, all of which are developed from the Transformer architecture~\cite{vaswani2017attention} initially proposed by Vaswani et al. in 2017. We consider adapting the pre-trained model to downstream tasks, such as text classification, summarizing, question answering, and sentiment analysis. Each downstream task is represented by a training dataset of input-output pairs: $\mathcal{D}: {(x_i, y_j)}_{i=1}^N,  _{j=1}^M$. Here, $x_i$ represents a sequence of input tokens. The output $y_j$ can vary depending on the task, for instance, it may be a sequence of tokens for text generation tasks, categorical labels for classification tasks, or continuous values for regression tasks. For example, for sentiment analysis, $x_i$ is a social media post, and $y_j$ is the categorical label; for the question-answering task, $x_i$ is the question, and $y_j$ is the answer; for the summarizing task, $x_i$ is the article and $y_j$ is the summary of the corresponding article. Consider $P_\theta (y,x)$ to be a text generation task-based pre-trained model. During the full fine-tuning, the model is initialized with its pre-trained weights $\theta_0$ and subsequently updated to $\theta_0 + \Delta \theta$. Here, $\Delta \theta$ represents the modifications made to the pre-trained weights, specifically tailored to enhance performance on the downstream task. These adjustments are derived by optimizing the following objective function:~\cite{hu2021lora}: 

\begin{equation}
    \max_{\theta} \sum_{(x,y) \in \mathcal{D}} \sum_{t=1}^{|y|} \log (P_{\Phi}(y_t \mid x, y_{<t}))
\end{equation}

Here, the objective function aims to maximize the cumulative log probability of correctly predicting each output token $y_t$, given the input sequence $x$ and all preceding output tokens.

A major drawback of full fine-tuning is that all the model parameters need to be trained for each downstream task, which makes the dimension of $|\Delta \theta|$ equal to $|\theta_0|$. In this paper, we introduce a parameter-efficient fine-tuning strategy that significantly decreases the number of trainable parameters from $\theta_0$ to $\phi$, where the dimension of $|\Delta \phi| \ll |\Delta \theta|$. 

\subsection{Proposed Method}

Transformer-based embedding models, such as BERT, depicted in Figure~\ref{fig:ttlora_architecture}(a), comprise multiple dense layers, including numerous encoder blocks and a fully connected feed-forward neural network. Each encoder block, detailed in Figure~\ref{fig:ttlora_architecture}(b), contains a complex sub-layer arrangement, specifically multi-head  attention layer and a fully connected feed-forward neural network. The weight matrices associated with these dense layers possess full rank, ensuring that each input feature uniquely contributes to the output and maximizes the learning capacity of the layer. However, when adapting to a specific downstream task, Aghajanyan et al.~\cite{aghajanyan2020intrinsic} show that the pre-trained LLMs have a low intrinsic dimension and Hu et al.~\cite{hu2021lora} hypothesize that the updates to the weights of the pre-trained models also have a low intrinsic rank, highlighting a potential area for model compression during adaptation to downstream tasks. Therefore, in this paper, for a pre-trained weight matrix $W_0 \in \mathbb{R}^{m \times n}$ and its corresponding update during adaptation $\Delta W \in \mathbb{R}^{m \times n}$, we constrain the update through the proposed TT-LoRA approach by presenting $\Delta W$ with low-rank representations, as shown in Figure~\ref{fig:ttlora_architecture}(c). 

To achieve a low-rank approximation of $\Delta W$, we utilize tensor decomposition~\cite{kolda2009tensor}, more specifically, Tensor Train (TT)\cite{12} decomposition. TT decomposition decomposes a tensor into a series of low-rank, small, three-dimensional tensors (cores). The product of these low-rank tensor cores provides an accurate approximation of the original tensor, significantly reducing its dimensionality while preserving essential structure and information. The fundamental property of TT decomposition is that each core tensor interacts only with its immediate predecessor and successor. This localized interaction allows operations, such as tensor core multiplication to approximate the original tensor, to be performed in a step-by-step manner, focusing on smaller, manageable pieces rather than the entire tensor at once. As a result, the complexity of tensor operations is significantly simplified, leading to substantial reductions in computational and memory requirements~\cite{oseledets2011tensor}. However, TT decomposition is particularly applicable to high-dimensional tensors. Therefore, in TT-LoRA, the matrix $\Delta W$, which is initialized using a random Gaussian distribution, is first represented as a $d$-dimensional tensor $\Delta \mathcal{W} \in \mathbb{R}^{k_1 \times \dots \times k_d}$. Here, $\prod_{i=1}^{d} k_i = m \times n$. The $d$-dimensional tensor $\Delta \mathcal{W}$ is then decomposed into $d$ number of small tensor cores $\mathcal{C}_1, \dots, \mathcal{C}_d$. The shape of each tensor core can be defined as $\mathcal{C}_i \in \mathbb{R}^{r_{i-1}, k_i, r_i}$, given the TT rank $[r_0, \dots , r_d]$, where the first ($r_0$) and last ($r_d$) TT ranks are 1.  Consequently, the total number of parameters in the tensor train decomposition of  $\Delta \mathcal{W}$ can be represented as:
\begin{equation}
    \Delta \mathcal{W} \approx \bigcup_{i=1}^d \mathcal{C}_i \in \mathbb{R}^{\sum_{i=1}^d r_{i-1} \times k_i \times r_i}
\end{equation}

Here, $r_{i-1} \times k_i \times r_i$ is the size of the $i$-th tensor core. During fine-tuning, the pre-trained weight matrix $W_0$ remains frozen and does not receive gradient updates, while the $\Delta \mathcal{W}$ contains trainable parameters.
The adapted weight matrix $W_{adapted}$ is given by:
\begin{equation}
    W_{adapted} = W_0 + \alpha(\Delta W)
\end{equation}
Using TT decomposition:
\begin{equation}
    W_{adapted} = W_0 + \alpha(\prod_{i=1}^{d} \mathcal{C}_i )
\end{equation}
Thus, this adapted layer applies a linear transformation to an input $x$ can be described as follows:

\begin{equation}
\textbf{}    \begin{aligned}
        y = W_0x + \alpha(\prod_{i=1}^{d} \mathcal{C}_i x)
    \end{aligned}
\end{equation}

Here, $W_0$ represents the pre-trained weight matrix of shape $m \times n$, $\Delta W$ is the update to the weight matrix in full fine-tuning of shape $m \times n$, $\Delta \mathcal{W}$ is the d-dimensional tensorized matrix of shape $k_1 \times \dots \times k_d$, $\mathcal{C}_i$ is the $i$-th tensor core of shape $r_{i-1} \times k_i \times r_i$ and $\alpha$ is a fixed scaling factor used to scale the update before adding to the pre-trained weight.

The compression ratio of TT-LoRA is closely related to the choice of TT ranks and the structure of d-dimensional tensor $\Delta \mathcal{W}$. For instance, consider a $W_0$ with dimensions $768 \times 2304$. In full fine-tuning, $\Delta W$ will have the exact dimensions of $768 \times 2304$, resulting in $1,769,472$ trainable parameters. However, with TT-LoRA, considering $\Delta \mathcal{W}$ as 7D tensor of shape $12 \times 8 \times 8 \times 3 \times 8 \times 8 \times 12$ with TT-rank of $5$ results in $1,135$ trainable parameters, achieving a remarkable $1560$x compression ratio. Reducing the TT-rank and/ or increasing the tensor dimension increases the compression ratio even further. To obtain the optimal TT-rank and tensor dimensions for our proposed method, we did a thorough hypermarameter search which is discussed in Section~\ref{hyper}.

%% file: 03-sec_results.tex
\begin{table*}[!th]
\begin{threeparttable}
\caption{Comparative analysis of various PEFT methods on the BERT family models \label{bert_results}}
\centering\footnotesize\setlength\tabcolsep{6pt}

    \renewcommand{\arraystretch}{0.95}
\begin{tabular}{l|c|c|c|c|c|c|c|c|c|c} 
 \hline
 \multirow{2}{*}{Model \& Method} & \multirow{2}{*}{\# Train. Param.} & \multirow{2}{*}{MNLI} & \multirow{2}{*}{SST-2} & \multirow{2}{*}{MRPC} & \multirow{2}{*}{CoLA} & \multirow{2}{*}{QNLI} & \multirow{2}{*}{QQP} & \multirow{2}{*}{RTE} & \multirow{2}{*}{STS-B} & \multirow{2}{*}{Avg.} \\  &&&&&&&&& \\ \hline
 DeBERTa-Base (FT)* & 139.19M & 88.67 & 94.61 & 91.98 & 59.32 & 93.04 & 91.42 & 68.23 & 91.10 & 84.79 \\
\hline
DeBERTa-Base (Adapters\(_{r=8}\))* & 0.94M & 87.69 & 94.72 & 88.88 & 54.19 & 92.95 & 85.52 & 59.20 & 89.68 & 81.60 \\
\hline
DeBERTa-Base (LoRA\(_{r=8}\))* & 0.30M & 87.30 & 94.95 & 92.84 & 60.56 & 93.35 & 85.19 & 80.14 & 90.13 & 85.56 \\
\hline
DeBERTa-Base (P-Tuning)* & 0.23M & 56.25 & 91.39 & 79.93 & 43.31 & 86.30 & 78.43 & 55.95 & 78.38 & 71.24 \\
\hline
DeBERTa-Base (LoRA\(_{r=4}\))* & 0.15M & 87.69 & 94.49 & 91.10 & 62.57 & 92.60 & 87.30 & 69.67 & 91.12 & 84.54 \\
\hline

DeBERTa-Base (Prefix)* & 0.15M & 60.32 & 88.87 & 81.22 & 45.82 & 83.28 & 82.22 & 59.57 & 84.99 & 73.28 \\
\hline
DeBERTa-Base (BitFit)* & 0.10M & 84.63 & 95.41 & 91.42 & 64.06 & 93.30 & 84.15 & 66.79 & 90.23 & 83.75 \\
\hline
DeBERTa-Base (LoRETTA\(_{adp}\))* & 0.10M & 85.93 & 95.30 & 93.53 & 60.84 & 92.99 & 84.08 & 75.50 & 91.32 & 84.96 \\
\hline
DeBERTa-Base (LoRETTA\(_{rep}\))* & 0.05M & 86.80 & 95.53 & 88.73 & 59.69 & 93.25 & 89.2 & 75.81 & 90.66 & 84.95 \\
\hline
DeBERTa-Base (Prompt)* & 0.01M & 77.63 & 92.43 & 81.90 & 32.99 & 80.30 & 78.15 & 62.81 & 56.71 & 70.36 \\
\hline
\textbf{DeBERTa-Base (TT-LoRA)} & \textbf{0.02M} &\textbf{83.1} & \textbf{94.15} & \textbf{90.68} & \textbf{70.26}&\textbf{91.01}&\textbf{85.57}& \textbf{75.09}& \textbf{90.57}& \textbf{85.05}
\\

\hline
\hline
\hline
RoBERTa-Base (FT)* &124M& 86.27 &93.46&88.97&74.20& 91.49&91.20&77.61&90.04 & 86.66\\
\hline

RoBERTa-Base (LoRA\(_{r=8}\))* & 0.63M & 86.82 & 94.01 & 91.48 & 62.08 & 92.39 & 85.71 & 74.51 & 90.48 & 84.69 \\
\hline
RoBERTa-Base (BitFit)* & 0.10M & 85.30 & 94.80 & 92.33 & 62.70 & 91.30 & 68.10 & 73.60 & 88.50 & 82.08 \\
\hline

RoBERTa-Base (LoRETTA\(_{adp}\))* & 0.10M & 85.61 & 94.38 & 91.08 & 62.70 & 92.12 & 87.22 & 78.70 & 90.26 & 85.26 \\
\hline
\textbf{RoBERTa-Base (TT-LoRA)} & \textbf{0.02M} & \textbf{85.76} & \textbf{93.57} & \textbf{87.05}  & \textbf{72.82} & \textbf{91.23} & \textbf{88.06} & \textbf{77.25} & \textbf{91.23} & \textbf{86.00}\\ 
\hline
\end{tabular}
\begin{tablenotes}
    \small
    \item Note: * represents results shown in previous work~\cite{yang2024loretta}.
\end{tablenotes}
\end{threeparttable}

\end{table*}

\begin{table*}[th!]
\begin{threeparttable}
    
\caption{Comparative analysis of various PEFT methods on the LLaMA family models \label{llama_results}}
\centering\footnotesize\setlength\tabcolsep{4pt}

    \renewcommand{\arraystretch}{0.95}
\begin{tabular}{|l|c|c|c|c|c|c|c|c|c|} 
 \hline
 Model \& &\multicolumn{7}{c|}{LLaMA2-7B}  &\multicolumn{2}{c|}{LLaMA3-8B}  \\  \cline{2-10} 
 Task & FT* & Adapter* & LoRA\(_{r=8}\)* & Prefix* & LoRETTA\(_{rep}\)* & LoRETTA\(_{adp}\)* & \textbf{TT-LoRA} & LoRA\(_{r=8}\) & \textbf{TT-LoRA} \\ \hline

 CB & 66.07 & 66.07 & 67.86 & 51.78 & 55.35 & 66.07 &  \textbf{85.71} & 71.43 & \textbf{85.71}\\
\hline
BoolQ & 84.6 & 71.8 & 84.8 & 78.6 & 78.1 & 87.0 & \textbf{86.78} & 88.93 & \textbf{88.13} \\
\hline
WSC & 63.46 & 62.50 & 62.50 & 61.53 & 57.61 & 63.46 & \textbf{67.30} & 63.46 & \textbf{66.35} \\
\hline

COPA & 86 & 84 & 81 & 83 & 86 & 87 & \textbf{81}  & 72 & \textbf{77.99} \\
\hline

Avg. & 75.03 & 71.09 & 74.04 & 68.72 & 69.26  & 75.88 & \textbf{80.19} & 73.96 & \textbf{79.55}\\\hline
 \#Train. Param. & 6738.42M& 50.33M& 4.19M& 1.31M& 0.51M & 0.88M & \textbf{0.1M} & 3.41M & \textbf{0.2M}  \\\hline




\end{tabular}
\begin{tablenotes}
    \small
    \item Note: * represents results shown in previous work~\cite{yang2024loretta}.
\end{tablenotes}
\end{threeparttable}
    \vspace{-4mm}
\end{table*}

We conduct experiments to evaluate the performance of TT-LoRA on various downstream tasks, ranging from natural language understanding (NLU) to generation (NLG), utilizing pre-trained LLMs of different scales. Specifically, we evaluate DeBERTa~\cite{he2020deberta} and RoBERTa~\cite{liu2019roberta} on Generalized Language Understanding Evaluation (GLUE)~\cite{wang2018glue} benchmark while utilizing SuperGLUE~\cite{wang2019superglue} benchmark on larger-scale models, such as LLaMA-2-7B~\cite{touvron2023llama} and LLaMA-3-8B~\cite{1}. In addition to reporting model performance on downstream tasks with TT-LoRA, we compare the results with baseline fine-tuning methods, such as Full fine-tuning (FT), Adapters~\cite{ding2023parameter}, Prompt tuning~\cite{lester2021power}, Prefix Tuning~\cite{li2021prefix}, P-tuning~\cite{liu2021p}, BitFit~\cite{zaken2021bitfit}, LoRA~\cite{hu2021lora}, and LoRETTA~\cite{yang2024loretta}. Furthermore, we conduct an extensive search for optimal parameters, including the best tensor shapes for $\Delta \mathcal{W}$ and the appropriate TT ranks. This effort aims to establish benchmarks that illustrate the trade-off between model compression and performance.

For this paper, our experiments utilized a system that integrates four NVIDIA Hopper (H100) GPUs, each paired with a corresponding NVIDIA Grace CPU via NVLink-C2C, facilitating rapid data transfer crucial for intensive computational tasks. The GPUs are equipped with 96GB of HBM2 memory, optimal for handling large models and datasets. 

\subsection{GLUE Experiments on BERT Family}

We initially conducted experiments on (RoBERTa) Robustly Optimized BERT Pretraining Approach~\cite{liu2019roberta}, which is an optimized method for training BERT (Bidirectional Encoder Representations from Transformers)~\cite{devlin2018bert}, a transformer-based LLM. Developed by researchers at Meta AI, RoBERTa revises BERT's pretraining methodology to improve the model's performance in several ways, such as dynamic masking, eliminating the next sentence prediction loss, and increasing the batch size while decreasing the learning rate. Apart from RoBERTa, We performed experiments on DeBERTa (Decoding-enhanced BERT with disentangled Attention)~\cite{he2020deberta}, a recent variant of BERT trained on a larger scale. Developed by Microsoft, DeBERTa improves the BERT architecture by introducing a novel disentangled attention mechanism that separately models the content and position, enhancing the model’s ability to understand contextual relationships in text. We utilized the pre-trained RoBERTa-base and DeBERTa-base from the HuggingFace Transformers library.

In RoBERTa and DeBERTa architecture, each encoder layer includes four weight matrices within the self-attention module ($W_q$, $W_k$, $W_v$, $W_o$) and a feed-forward neural network (FFNN). While TT-LoRA can be applied to any of the weight matrices in a neural network to reduce the number of trainable parameters, our experiments specifically target $W_q$ and $W_v$. This focus aligns with findings from the LoRA paper, which indicates that models achieve optimal performance when these particular weights are fine-tuned, while the remaining weights are kept frozen~\cite{hu2021lora}. We performed hyperparameter optimization by fine-tuning the model with different TT-LoRA parameter initializations, the details of which will be presented later. Using the HyperBand optimizer~\cite{li2017hyperband}, we identified the most efficient parameters. For each run, the model was fine-tuned for up to 20 epochs with an early stopping criterion of 5 epochs. Specifically, the training was halted if the validation loss did not improve for 5 consecutive epochs. The best model was selected based on the lowest observed validation loss from these runs.For reporting performance on the GLUE benchmark tasks, we use the following metrics: matched accuracy for MNLI, Matthews correlation coefficient for CoLA, Spearman correlation coefficient for STS-B, F1 score for both MRPC and QQP, and accuracy for all other tasks. Table~\ref{bert_results} summarizes the downstream task performance comparison between TT-LoRA and other baseline PEFT methods. 

As shown in Table~\ref{bert_results}, TT-LoRA consistently achieves superior or comparable performance to other PEFT methods when FT DeBERTa on GLUE tasks, with no more than 0.2M trainable parameters. Consequently, TT-LoRA stands out for its efficiency by outperforming 9 out of 10 FT approaches in both model compression and accuracy.Prompt Tuning, which utilizes just 0.01M trainable parameters compared to TT-LoRA's 0.02M, surpasses TT-LoRA in model compression but significantly suffers in performance. Nevertheless, TT-LoRA requires merely twice the trainable parameters of Prompt Tuning while achieving, on average, about 15\% higher model accuracy. TT-LoRA also achieves achieves substantial model compression when FT RoBERTa, significantly reducing the number of trainable parameters. Specifically, TT-LoRA has reduced trainable parameters by approximately factors of $6,200 \times$ for full FT, $31.5 \times$ for LoRA, $5 \times$ for BitFit, and $5 \times$ for LoRETTA$_{adp}$. Despite this reduction, TT-LoRA outperforms other PEFT methods, except for full FT, in average model accuracy across various GLUE task sets.

\subsection{SuperGLUE Experiments on LLaMA Family}
Encouraged by the outcomes observed with DeBERTa and RoBERTa models, we extended our experiments to include the larger-scale Large Language Model at Meta AI (LLaMA)~\cite{llama} models. LLaMA is a series of larger-scale language models designed for various natural language understanding and generation tasks. In our experiments, we utilized LLaMA-2-7B and LLaMA-3-8B models to assess the effectiveness of our proposed PEFT method. We chose these LLaMA models due to their extensive parameter sets, ranging into the billions, which provide a rigorous test environment to evaluate how well our PEFT approach can compress and optimize these substantial models while maintaining comparable performance levels. For our experiments, we utilized the pre-trained LLaMA2-7b and LLaMA3-8b models available in HuggingFace Transformers library. We conducted a comparative analysis of TT-LoRA against other baseline PEFT methods using the SuperGLUE benchmark tasks and the results are summarized in Table~\ref{llama_results}. Similar to BERT family experiments, TT-LoRA has been applied to $W_q$ and $W_v$ weight matrices of the self-attention module of LLaMA models and has been trained over multiple epochs, stopping the process if the validation loss did not improve for 5 consecutive epochs. The best model is then chosen from these runs based on the lowest observed validation loss. For reporting performance on the SuperGLUE benchmark tasks, we used F1 score for both CB and WSC tasks, and accuracy for both BoolQ and COPA tasks. 

Table~\ref{llama_results} highlights TT-LoRA's performance on the LLaMA2-7B model, where it consistently outperforms all other PEFT methods on CB and WSC tasks and all but LoRETTA$_{rep}$ on the BoolQ task. It achieves comparable performance on the COPA task and, on average, surpasses all competing PEFT methods. This higher model performance is achieved while attaining significant reductions in trainable parameters compared to our baselines by factors of approximately 67,384x (full fine-tuning, FT), 500x (Adapter), 41.9x (LoRA$_{r=8}$), 13.1x (Prefix), 5.1x (LoRETTA$_{rep}$), and 8.8x (LoRETTA$_{adp}$). 

We extended our experiment by FT the LLaMA3-8B model on the SuperGLUE benchmark and compared its performance against LoRA. As shown in Table~\ref{llama_results}, TT-LoRA either matched or exceeded the performance of LoRA across all tasks while achieving a remarkable reduction in trainable parameters, which is approximately $170.5 \times$ fewer parameters compared to LoRA.

\subsection{Memory Performance}

\begin{figure}[t!]
    \centering
    \includegraphics[width=0.4\textwidth]{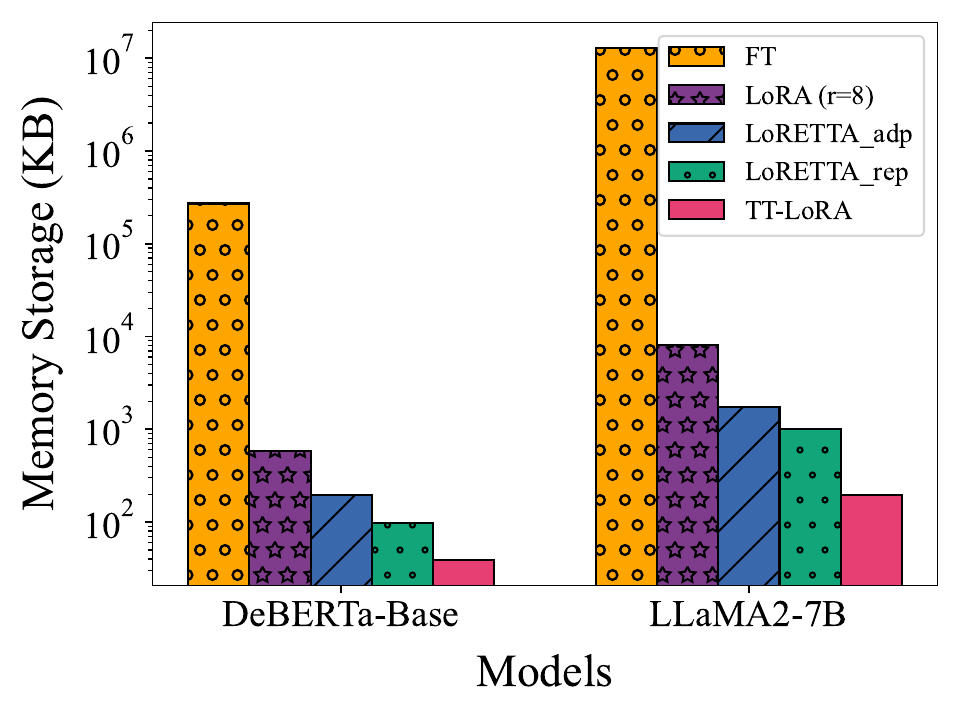}
    \caption{Memory storage comparison for trainable parameters of various PEFT methods and LLMs.}
    \label{fig:memory}
\end{figure}

We evaluate the storage requirements of various PEFT methods applied to fine-tune the DeBERTa and LLaMA2-7B models. Specifically, we select the three top-performing PEFT methods from Table~\ref{bert_results} and Table~\ref{llama_results}. The storage calculations are based on the assumption that the model weights are stored with 16-bit precision. As demonstrated in Figure~\ref{fig:memory}, TT-LoRA  reduces the storage requirements for trainable parameters when used with DeBERTa, necessitating only $39$ KB. This represents a significant reduction in storage needs, achieving reductions by factors of approximately $2.5 \times$, $5 \times$, $15 \times$, and $7000 \times$ compared to LoRETTA$_{rep}$, LoRETTA$_{adp}$, LoRA$_{r=8}$, and full FT, respectively. A similar memory efficiency is achieved through TT-LoRA when used with LLaMA2-7B, requiring approximately $195$ KB of storage for trainable parameters. This storage efficiency significantly surpasses that of other PEFT methods, compared to our baselines reducing storage demands by factors of roughly $5 \times$ (LoRETTA$_{rep}$), $8.8 \times$ (LoRETTA$_{adp}$), $42 \times$ (LoRA$_{r=8}$), and $65,925 \times$ (full fine-tuning, FT). The reduced storage requirement of TT-LoRA establishes it as a highly efficient approach for FT LLMs, particularly on hardware with limited resources. This efficiency facilitates broader deployment options and ensures that larger-scale LLM capabilities are accessible even in constrained environments.

\subsection{Hyperparameters}
\label{hyper}

\begin{table}
    \caption{Hyperparameter Search Used for Experiments}
\centering\footnotesize\setlength\tabcolsep{4pt}
    \renewcommand{\arraystretch}{0.95}
    \begin{tabular}{|l|l|l|}
    \hline
         \multirow{18}{*}{Tensor Shape}&  \multirow{6}{*}{DeBERTa}& [64, 36, 12, 64] \\ \cline{3-3} && [32, 12, 3, 4, 12, 32] \\ \cline{3-3}&& [12, 8, 8, 3, 8 , 8, 12] \\ \cline{3-3}&&[32, 16, 2, 3, 3, 3, 2, 32] \\ \cline{3-3}&&[32, 4, 2, 2, 2, 3, 3, 3, 2, 32] \\ \cline{3-3}&&[16, 2, 4, 2, 2, 2, 3, 3, 3, 2, 2, 16] \\ \cline{2-3}
         
         &\multirow{6}{*}{RoBERTa}& [64, 16, 9, 64] \\ \cline{3-3} && [12, 8, 8, 8, 8, 12] \\ \cline{3-3}&& [12, 8, 8, 2, 4, 8, 12] \\ \cline{3-3}&&[12, 8, 8, 2, 2, 2, 8, 12] \\ \cline{3-3}&&[8, 6, 2, 2, 4, 4, 2, 2, 6, 8] \\ \cline{3-3}&&[8, 6, 2, 2, 2, 2, 2, 2, 2, 2, 6, 8] \\ \cline{2-3}
         
         &\multirow{5}{*}{LLaMA2-7B}& [128, 32, 32, 128] \\ \cline{3-3} && [16, 16, 16, 16, 16, 16] \\ \cline{3-3}&& [16, 16, 16, 4, 4, 16, 16] \\ \cline{3-3}&LLaMA3-8B&[16, 16, 4, 4, 4, 4, 16, 16] \\ \cline{3-3}&&[16, 8, 4, 4, 2, 2, 4, 4, 8, 16] \\ \cline{3-3}&&[16, 4, 4, 4, 2, 2, 2, 2, 4, 4, 4, 16] \\ \hline

         \multirow{4}{*}{Ranks}&  DeBERTa & \multirow{2}{*}{5, 8, 10, 12, 16}\\ \cline{2-2} &RoBERTa &\\\cline{2-3} & 
         LLaMA2-7B & \multirow{2}{*}{1, 2, 4, 5, 8, 10, 12, 16}\\ \cline{2-2} &LLaMA3-8B &\\\hline
         Alpha&  \multicolumn{2}{c|}{1, 2, 4, 8, 10, 12, 16, 32}\\ \hline
         Learning Rate&  \multicolumn{2}{c|}{$1e-5$, $1e-4$, $5e-5$, $5e-4$}\\ \hline
    \end{tabular}
    \vspace{-5mm}
    \label{tab:hyper}
\end{table}

\begin{figure}[htb]
    \centering
    \includegraphics[width=0.5\textwidth]{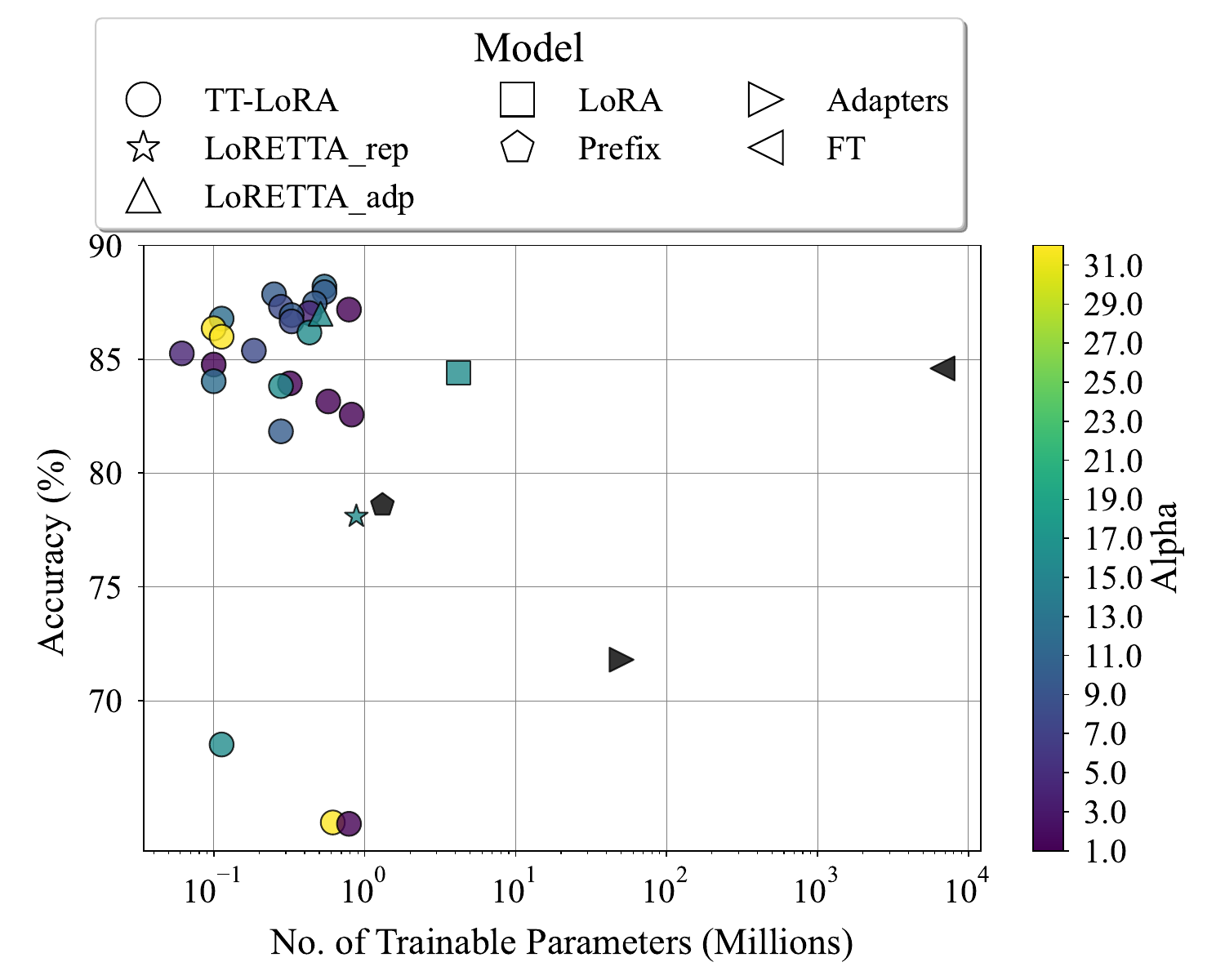}
    \caption{Model compression vs. accuracy of LLaMA2-7B model with TT-LoRA having different tensor format and other PEFT approaches on SuperGLUE BoolQ benchmark.}
    \label{fig:boolq}
\end{figure}

To optimize the performance of TT-LoRA, we conducted a comprehensive hyperparameter search using Ray Tune~\cite{ray_tune}, a scalable hyperparameter optimization framework. The hyperparameters are selected to ensure a robust comparison of model performance across different configurations and tasks. We employed the HyperBand optimizer~\cite{li2017hyperband} to explore a discrete search space for hyperparameter combinations, aiming to minimize validation loss with fewer simulations. This approach allowed us to estimate the optimal parameters efficiently, performing around 200 searches. The search space included learning rates and specific model-related parameters such as tensor shapes, ranks, and alpha values, as highlighted in Table~\ref{tab:hyper}. The best model configurations were determined based on the lowest validation loss observed during the tuning Ray Tune trials. The optimal hyperparameters for TT-LoRA, such as tensor shape, tensor rank, $\alpha$, and learning rate are summarized in Table~\ref{tab:opt_hyp}. The hyperparameters used for other PEFT methods are demonstrated in~\cite{yang2024loretta}.

\begin{table}
    \caption{Optimal Hyperparameters for TT-LoRA}
    \centering
    \renewcommand{\arraystretch}{0.92}
    \begin{tabular}{|c|c|c|}
    \hline
      Model & Hyperparameter & Value\\\hline
         \multirow{4}{*}{DeBERTa} & Tensor Shape & [64, 36, 12, 64]\\ \cline{2-3}
         & Tensor Rank & 5\\ \cline{2-3}
         & $\alpha$ & 8\\ \cline{2-3}
         & Learning Rate & $1e-5$\\ \hline
         \multirow{4}{*}{RoBERTa} & Tensor Shape & [64, 16, 9, 64]\\ \cline{2-3}
         & Tensor Rank & 4\\ \cline{2-3}
         & $\alpha$ & 1, 2, 8, 10\\ \cline{2-3}
         & Learning Rate & $1e-1$\\ \hline
         \multirow{4}{*}{LLaMA2-7B} & Tensor Shape & [16, 16, 16, 16, 16, 16]\\ \cline{2-3}
         & Tensor Rank & 5\\ \cline{2-3}
         & $\alpha$ & 4\\ \cline{2-3}
         & Learning Rate & $1e-5$\\ \hline

         \multirow{4}{*}{LLaMA3-8B} & Tensor Shape & [16, 4, 4, 4, 2, 2, 2, 2, 4, 4, 4, 16]\\ \cline{2-3}
         & Tensor Rank & 10\\ \cline{2-3}
         & $\alpha$ & 12\\ \cline{2-3}
         & Learning Rate & $1e-1$\\ \hline
    \end{tabular}
        \vspace{-5mm}
    \label{tab:opt_hyp}
\end{table}

In addition to identifying the optimal model configuration, the logs generated by Ray Tune provide a detailed analysis of the relationship between model performance and the number of trainable parameters. This insight is crucial for optimizing a PEFT method, especially in resource-constrained environments where efficiency is also important. Figure~\ref{fig:boolq} illustrates how the number of trainable parameters, determined by specific tensor shapes and ranks, affects the accuracy of LLaMA2-7B model fine-tuned with TT-LoRA. The heatmap in Figure~\ref{fig:boolq} depicts the $\alpha$ value, signifying the magnitude of updates to the pre-trained weights during fine-tuning. As shown in Figure~\ref{fig:boolq}, model accuracy tends to decrease as the number of trainable parameters is reduced. In addition to model parameters, the value of $\alpha$ also influences the model accuracy. Nonetheless, TT-LoRA maintains superior or comparable performance to other PEFT methods, even with fewer parameters, demonstrating its capability to effectively minimize trainable parameters within strict hardware limits.


%% file: 04-sec_conclusion.tex
We introduced TT-LoRA, a parameter-efficient fine-tuning approach that leverages tensor train decomposition to significantly reduce the number of trainable parameters. TT-LoRA demonstrated substantial model compression and improved performance when fine-tuning BERT and LLaMA-based models across various tasks. Compared to other state-of-the-art PEFT methods, TT-LoRA achieved higher or comparable average accuracy with a significantly smaller model size, underscoring its effectiveness in both parameter reduction and performance enhancement.
For future work, we aim to extend TT-LoRA to compress larger-scale models such as LLaMA3.1-405B, Grok 2.0, and Mistral Large. Additionally, we plan to explore the compression of additional layers within LLMs to achieve even greater levels of compression.